\documentclass{article}
\usepackage{spconf,amsmath,graphicx,amssymb}
\usepackage{algorithmic,algorithm,color}

\usepackage{fancyheadings}
\newcommand{\argmin}{\arg\!\min}

\title{Tumour ROI Estimation in Ultrasound Images via Radon Barcodes\\ in Patients with Locally Advanced Breast Cancer}

\name{Hamid R. Tizhoosh$^{1}$, Mehrdad J. Gangeh$^{*2,~3}$, Hadi Tadayyon$^{2,~3}$, Gregory J. Czarnota$^{2,~3}$}

\address{$^{1}$  KIMIA Lab, University of Waterloo, Canada, tizhoosh@uwaterloo.ca \\
		$^{2}$ Depts. of Medical Biophysics and Radiation Oncology, University of Toronto, Canada \\
     $^{3}$ Dept. of Radiation Oncology and Imaging Research, Sunnybrook Health Sciences Center, Canada \\
      \{$^*$mehrdad.gangeh,hadi.tadayyon,gregory.czarnota\}@sunnybrook.ca}	

\begin{document}
%
\maketitle
\begin{abstract}
Quantitative ultrasound (QUS) methods provide a promising framework that can non-invasively and inexpensively be used to predict or assess the tumour response to cancer treatment. The first step in using the QUS methods is to select a region of interest (ROI) inside the tumour in ultrasound images. Manual segmentation, however, is very time consuming and tedious. In this paper, a semi-automated approach will be proposed to roughly localize an ROI for a tumour in ultrasound images of patients with locally advanced breast cancer (LABC). Content-based barcodes, a recently introduced binary descriptor based on Radon transform, were used in order to find similar cases and estimate a bounding box surrounding the tumour. Experiments with 33 B-scan images resulted in promising results with an accuracy of $81\%$.
\end{abstract}
\section{Introduction}
\label{sec:intro}
Therapeutic cancer response monitoring has long been facilitated using several functional imaging modalities~\cite{MI:Brindle08} such as positron emission tomography (PET)~\cite{MI:Juweid06}, magnetic resonance imaging (MRI)~\cite{MI:Witney10}, and diffuse optical imaging (DOI)~\cite{MI:Soliman10}. Although these imaging technologies can potentially provide an early assessment of cell death at microscopic level~\cite{MI:Brindle08}, they suffer from two main drawbacks: being expensive, and requiring an exogenous agent, which is also costly and may cause some side effects and allergic reactions~\cite{MI:Czarnota10}. 
In comparison, techniques based on quantitative ultrasound (QUS)~\cite{MI:Czarnota99,MI:AliSN15} facilitate a promising cost-effective, non-invasive, and rapid framework for the early assessment of cancer therapy effects using ultrasound. Moreover, QUS methods rely on endogenous contrast -- generated by the very process of cell death itself -- to evaluate treatment effectiveness, which alleviates the requirement for injecting external contrast agents.

In addition to the recent advances in cancer response monitoring using QUS methods~\cite{Gangeh:TMI14,Gangeh:TMI15,Gangeh:SPIE14}, the applications of these techniques have recently been extended to treatment prediction~\cite{MI:Tadayyon15} and tissue characterization using 3-D automated breast ultrasound (ABUS) technology~\cite{Gangeh:SPIE16}.

Irrespective of the applications for which the QUS methods are to be used, no matter whether it is response monitoring, response prediction, or tissue characterization, the first step is to identify the frames containing the tumour, and subsequently contouring a region of interest (ROI) inside the tumour for further analysis such as the computation of parametric maps using spectroscopy methods~\cite{MI:Lizzi97}, etc. This step has to be done on ultrasound B-mode images and since currently, there is no automated software to contour inside the tumours, performing this initial step manually, is very tedious, time consuming, and accounts for a huge amount of efforts placed in each research project related to QUS methods. In order to have an idea on the timing required for this step, we highlight that for a typical preclinical study involving 100 mice, even if we select ROIs from only 10 frames in ``pre-treatment'' and another 10 from ``post-treatment'' scans, $100\times20 = 2000$ ROIs should be contoured that may take days to weeks to complete depending to the speed and experience of the technician. The problem is even more severe when dealing with 3-D ultrasound scanners such as ABUS systems when tens of ROIs should be contoured for each patient particularly for those with large tumour size, which is usually the case, for example, in aggressive types of cancer tumours such as locally advanced breast cancer (LABC)~\cite{LABC:Giordano03}.

This research addresses the aforementioned problem, to some extent, by proposing a novel semi-automated tumour localization approach for ROI estimation in patients with LABC. The ultrasound B-mode images were acquired from LABC patients before treatment onset and the ultimate goal was to use the automated selected ROIs for cancer response prediction~\cite{MI:Tadayyon15}. This paper, however, only focuses on the methods and results for a proposed barcode approach for rough tumour localization. Content-based barcodes represent a novel class of binary descriptors to tag digital images \cite{Tizhoosh2015}. The ultimate goal is to find similar images when a query image is provided by the user. In this paper, instead of extracting the contour of the tumour, we estimate the coordinates of a bounding box which surrounds the tumour by finding similar images, and by building a weighted average of their bounding boxes.

\section{Patient Data}

Patient data collection, from ten LABC patients with tumour sizes between 5 and 15~cm, was performed in accordance with the clinical research ethics approved by Sunnybrook Health Sciences Centre. Cancer diagnosis was confirmed via biopsy on all patients and magnetic resonance imaging (MRI) was performed in order to measure the size of tumour. Ultrasound (US) data was acquired from all patients before the start of neoadjuvant chemotherapy (``pre-treatment''). The US data acquisition was carried out using a Sonix RP ultrasound system, equipped with an L14-5/60 linear transducer with a centre frequency of $\sim$7~MHz. The transducer's focus was set at the midline of the tumour with a maximum depth of 4-6~cm, depending on tumour size and location. Three to five scan planes were obtained from the tumour, depending on its size, with a scan plane separation of $\sim$1~cm.

\section{Segmentation Approach}

As evident from the relatively low contrast images shown in Fig.~\ref{fig:sampleImages}, any segmentation technique will have difficulty to extract the tumours as marked by the expert. This was verified by experimenting using several methods such as thresholding, active contours, and watershed segmentation. As the ultimate goal is to characterize the tumour for treatment prediction, we can aim to get an ROI segmented that roughly sketches the tumour region instead of attempting to extract the precise tumour contour. Hence, we propose to use a novel approach, namely barcode-guided \emph{ROI segmentation}, which aims at finding a bounding box around the tumour instead of actual tumour contour.

The proposed approach indexed all available ground-truth images first by assigning two barcodes to each bounding box (of each ground-truth): a ``global'' barcode for the entire image, and a ``local'' (ROI-based) barcode for the bounding box. Similar to atlas-based methods, segmentation (in this case, ROI estimation) was subsequently performed through finding similar cases in the database. As for a query (new) image, a fixed-size ROI was first defined by asking the user to provide a mouse click in the centre of the tumour. The query image was subsequently tagged with two barcodes (global and ROI-based). By comparing the bar codes computed for the query image with those in the training set using a similarity measure, the top most similar tumours were identified and used to estimate the location of the tumour in the query image.

\subsection{Barcodes for Rough Localization}

The notion of Radon barcodes for image retrieval was first introduced by Tizhoosh~\cite{Tizhoosh2015}. The literature on medical image retrieval is vast. Ghosh \emph{et al.}~\cite{ghosh2011} have reviewed on-line systems for content-based medical image retrieval such as GoldMiner, BioText, FigureSearch, Yottalook, Yale Image Finder, IRMA, and iMedline. Multiple surveys are now available that review recent literature in content-based image retrieval (CBIR)~\cite{shandilya2010,rajam2013,dharani2013}. Recently \emph{autoencoders} have also been employed for image retrieval \cite{Tizhoosh2015arxiv}.

\begin{figure}[tbp]
\begin{center}
\includegraphics[width=0.70\columnwidth]{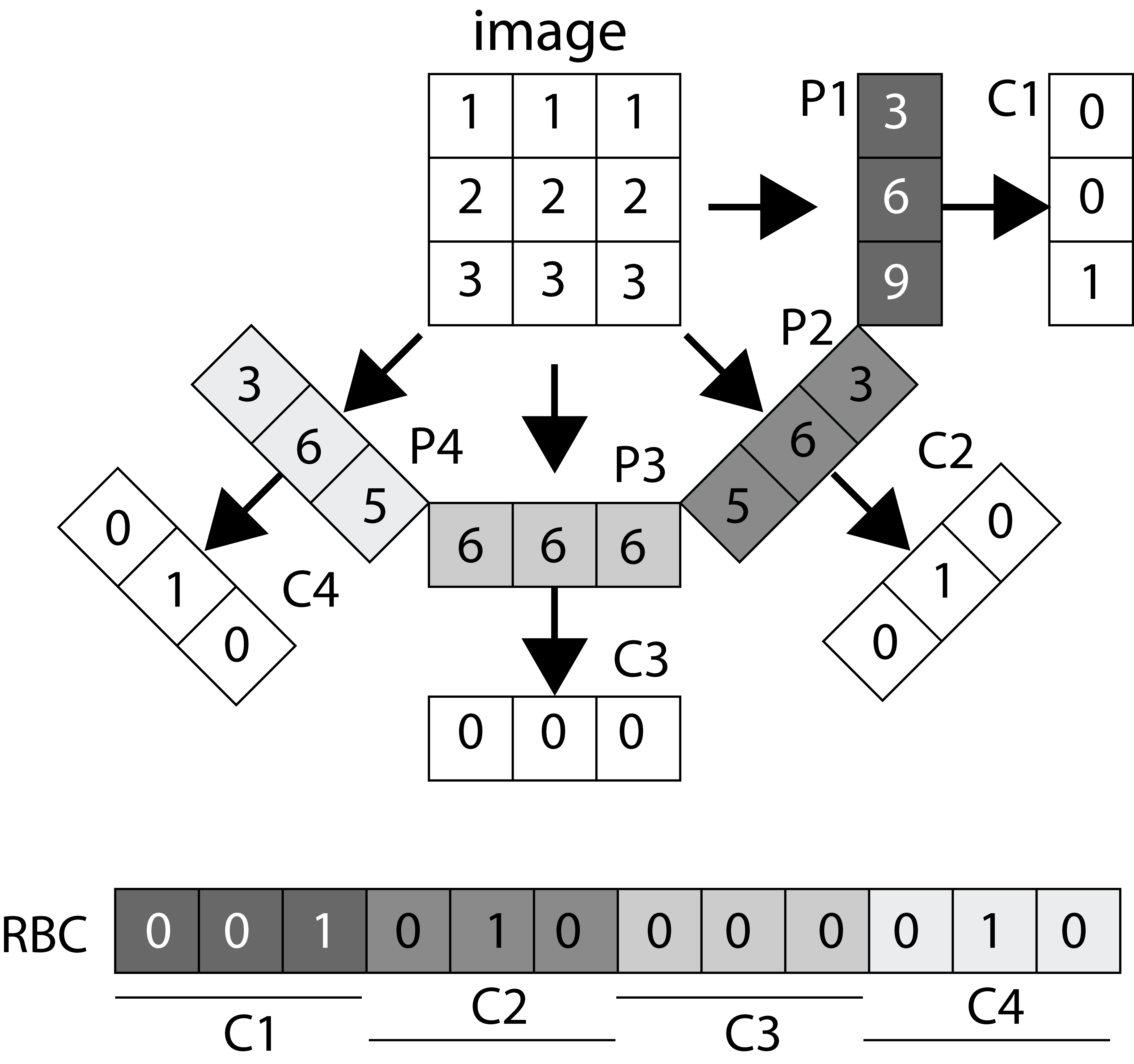}
\caption{Radon projections (P1, P2, P3, P4) are thresholded to generate code fragments (C1, C2, C3, C4). The concatenation of all code fragments delivers the barcode RBC. }
\label{fig:RBC}
\end{center}
\end{figure}

Considering an image as a 2-D function $I(x,y)$, it can be projected along a number of projection angles, which is the sum of $I(x,y)$ values along lines created for each angle $\theta$. The projection creates a new image $R(\rho,\theta)$ with $\rho = x \cos \theta + y \sin \theta$. Using the Dirac delta function $\delta(\cdot)$ the Radon transform can be written as
\begin{equation}
R(\rho,\theta) = \int\limits_{-\infty}^{+\infty} \int\limits_{-\infty}^{+\infty} I(x,y) \delta(\rho-x\cos \theta-y\sin\theta) dx dy.
\end{equation}
It has recently been proposed to threshold all projections for individual angles based on a ``local'' threshold for that angle to create a barcode of all thresholded projections \cite{Tizhoosh2015}: ``A simple way for thresholding the projections is to calculate a typical value via median operator applied on all non-zero values of each projection''.  Fig.~\ref{fig:RBC} illustrates how Radon barcodes (RBCs) are generated.

\subsection{Local versus Global Barcodes}

\begin{figure}[t]
\begin{center}
\includegraphics[width=0.6\columnwidth]{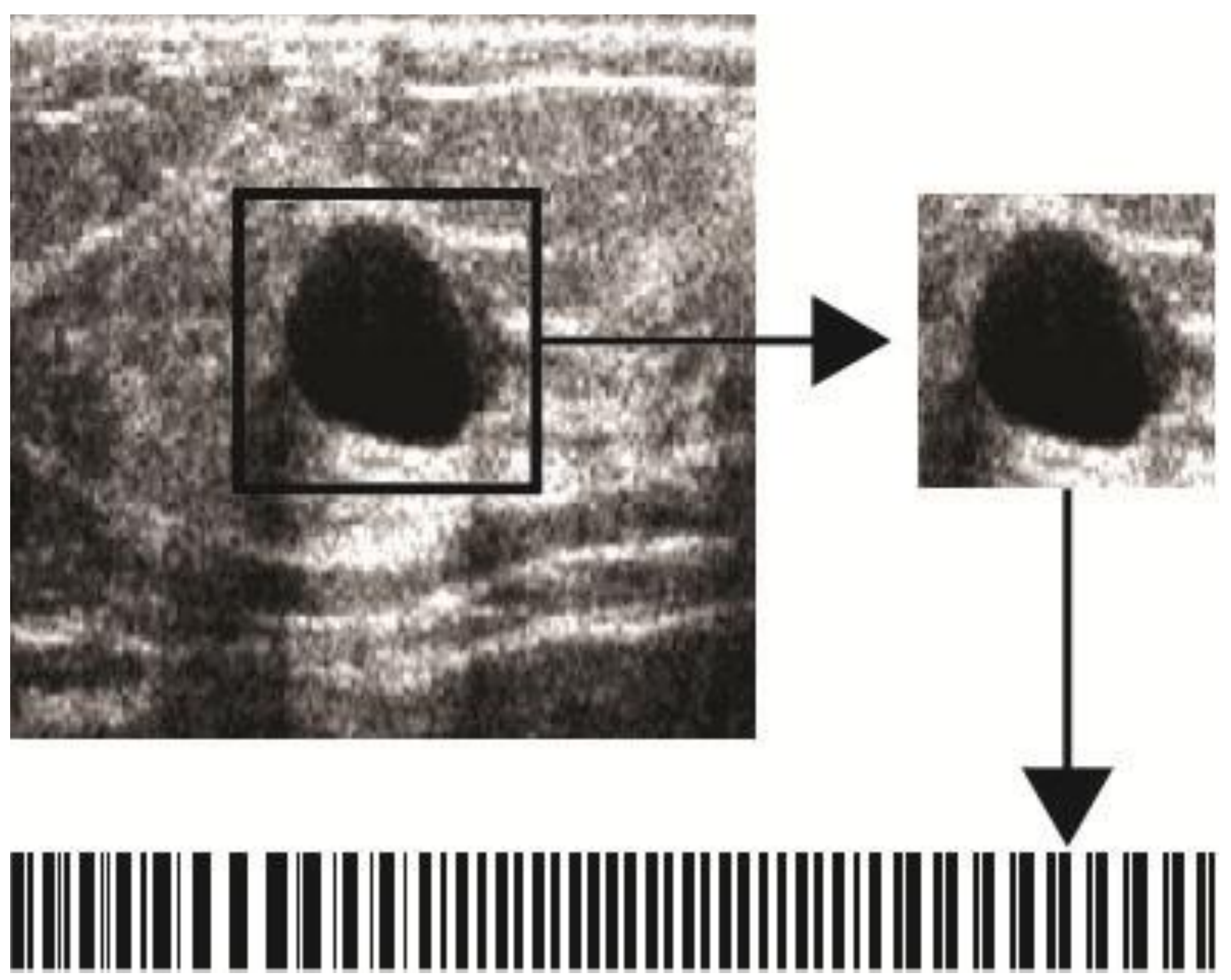}
\caption{Generation of a barcode for an ROI, here a breast lesion in an ultrasound scan (source: \cite{Tizhoosh2015}).}
\label{fig:ROIselection}
\end{center}
\end{figure}

In \cite{Tizhoosh2015}, only ``global'' barcodes were used, meaning that one barcode was extracted for the entire image. But it was recognized that using ``local'' barcodes may be of more significance when dealing with specific regions of interest (ROIs) (see Fig.~\ref{fig:ROIselection}). In this research, the two were combined and every image was tagged with two barcodes. The ``global'' barcode captures the general appearance of the image, and the ``local'' (or ROI-based) barcode captures the texture and intensity variations of the tumour.

\begin{figure}[h]
\begin{center}
\includegraphics[width=0.8\columnwidth]{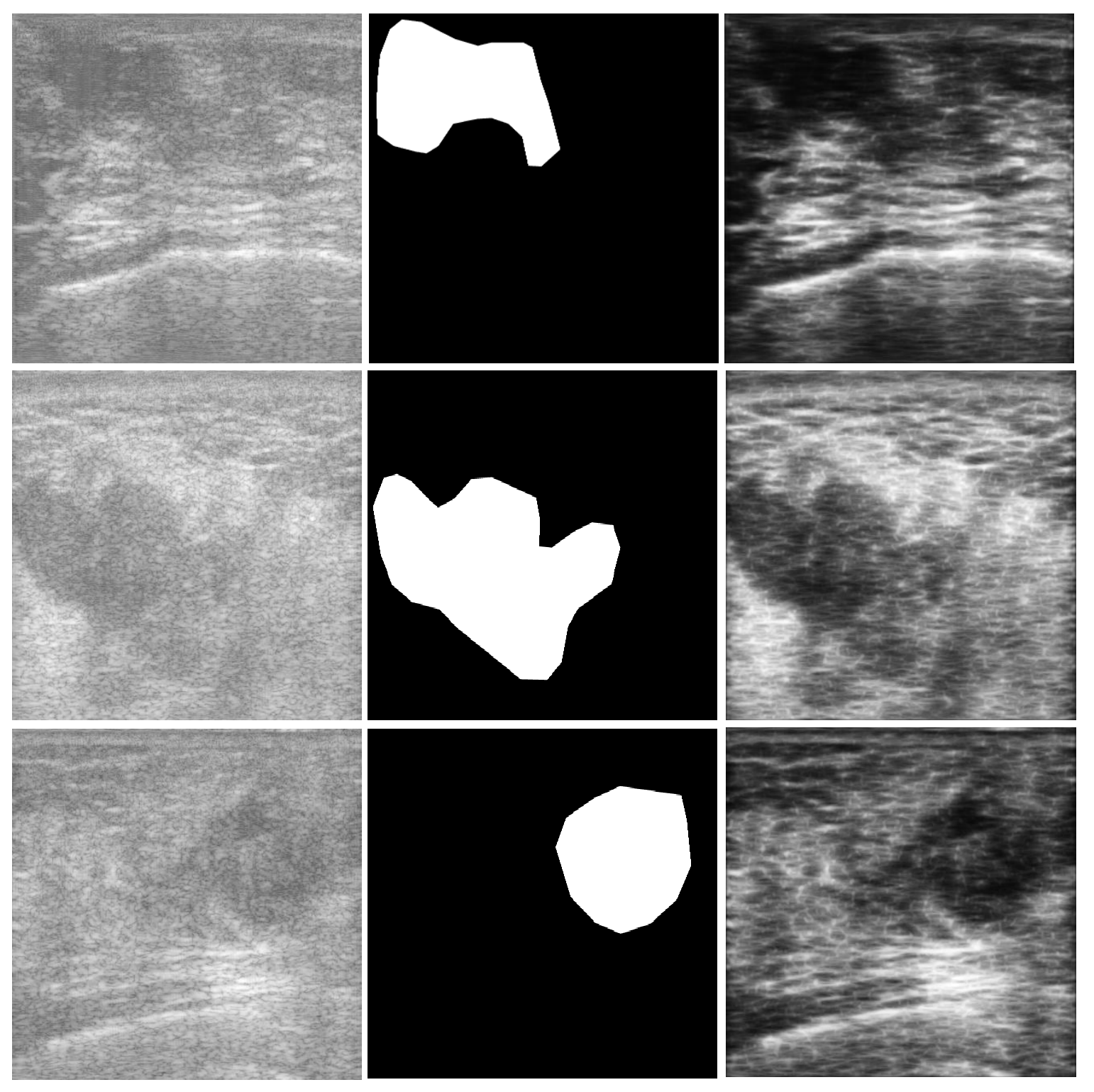}
\caption{Sample images (\emph{left}) with their ground-truth marked by an expert (\emph{middle}) and enhanced version via hyperbolization and noise filtering (\emph{right}). }
\label{fig:sampleImages}
\end{center}
\end{figure}

\textbf{Preprocessing of Images --} Before calculating content-based barcodes, the quality of each image was enhanced by modifying its contrast. For this purpose, a fuzzy hyperbolization~\cite{Tizhoosh1995,Tizhoosh1997} was employed that modified all gray-levels $g\in\{0,1,2,\dots,L-1\}$ ($L=$ max. number of gray-levels) to generate new gray-levels $g'$:
\begin{equation}
g'= (L-1)/(e^{-1}-1) \left[ e^{-\mu(g)^\beta}-1\right],
\end{equation}
where $\mu(g)\in[0,1]$ was a proper membership function, and $\beta>1$ had darkening effect on the image.
To suppress the speckle noise present in ultrasound images, sticks filter~\cite{Czerwinski1998} was used. Fig.~\ref{fig:sampleImages} shows some sample images with their corresponding ground-truth, and preprocessed versions.

\subsection{Barcode-Guided ROI Estimation}

\begin{figure}[t]
\begin{center}
\includegraphics[height=1.4in, width=1.4in]{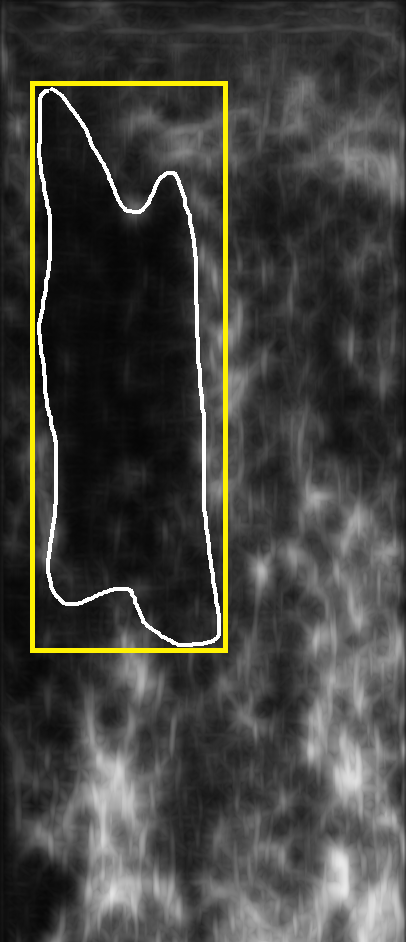}
\caption{Bounding Box for a ground-truth contour.}
\label{fig:BoundingBox}
\end{center}
\end{figure}

\begin{figure}[t]
\begin{center}
\includegraphics[width=0.74\columnwidth]{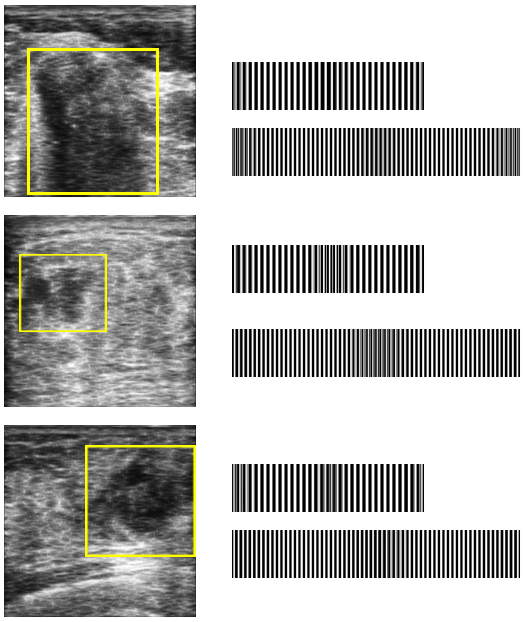}
\caption{Sample barcodes: For each image, two barcodes are extracted, one for the entire image (\emph{with more bits}), and one for the bounding box around the tumour (\emph{shorter barcode}). }
\label{fig:sampleBarCodes}
\end{center}
\end{figure}

Assuming that there is a set of training images along with their ground-truth segments, we propose to use barcodes in order to estimate the location of a bounding box $\mathbf{b}=(x_s,y_s;x_e,y_e)$ around the tumour (contoured by the expert as a ground-truth) with starting coordinates $(x_s,y_s)$ and ending coordinates $(x_e,y_e)$ (Fig. \ref{fig:BoundingBox}). Two barcodes were assigned to each bounding box: a global barcode for the entire image, and a local barcode for the bounding box (Fig. \ref{fig:sampleBarCodes}). The rough localization of a bounding box for the tumour can be formulated as a search problem: given a database consisting of $N$ training images $I_1,I_2,\dots,I_N$ with their corresponding bounding boxes $\mathbf{b}_1,\mathbf{b}_2,\dots,\mathbf{b}_N$, and global and ROI Radon barcodes $(\mathbf{r}_1^{\textup{G}},\mathbf{r}_1^{\textup{ROI}}),(\mathbf{r}_2^{\textup{G}},\mathbf{r}_2^{\textup{ROI}}),\dots,(\mathbf{r}_N^{\textup{G}},\mathbf{r}_N^{\textup{ROI}})$, the bounding box $\mathbf{b}_{\textup{query}}$ was formulated for a query image $I_{\textup{query}}$ by first finding the top $M$ similar images in the database via Hamming distance between corresponding barcodes:
\begin{equation}
\argmin_{i=1,2,\dots,N} \left(\textrm{xor}(\mathbf{r}_i^{\textup{G}},\mathbf{r}_{\textup{query}}^{\textup{G}}) + \textrm{xor}(\mathbf{r}_i^{\textup{ROI}},\mathbf{r}_{\textup{query}}^{\textup{ROI}})\right)
\end{equation}
After finding the dissimilarities for all cases, the top $M$ results were selected by sorting the list in decreasing order in a vector $\mathbf{v}$. Eventually, an ROI was estimated for the query image using:
\begin{equation}
\label{eq:estimate}
\mathbf{b}_{\textup{\textup{query}}}(l) = \left(\sum\limits_{i=1}^M \mathbf{b}_{v(i)}(l) w(i) \right)/\sum\limits_{i=1}^M w(i),
\end{equation}
where $l=1,2,3,4$ corresponded to $x_s,y_s,x_e$, and $y_e$ for each bounding box. The weights were calculated as follows:
\begin{equation}
\label{eq:weights}
w(i) = 1 - \mathbf{v}(i) /\max_j \mathbf{v}(j).
\end{equation}
Both query bounding box and the estimated bounding box were then reshaped into binary images $B_q$ and $B_e$ and compared using Dice coefficient $D$ to measure the accuracy: $D = {2 |B_q \cup B_e|}/({|B_q| + |B_e|})$. Algorithm \ref{alg:BarcodeROIEtimation} provides the details of the proposed method.

\begin{figure}[tb]
\begin{center}
\includegraphics[height=1in, width=1in]{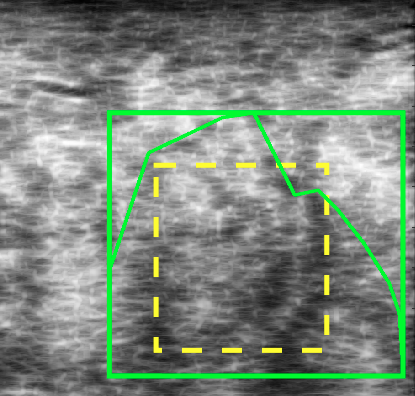}
\includegraphics[height=1in, width=1in]{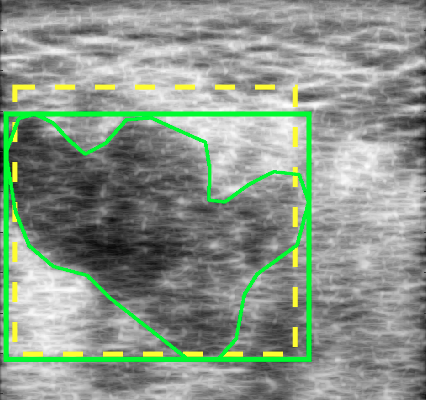}
\includegraphics[height=1in, width=1in]{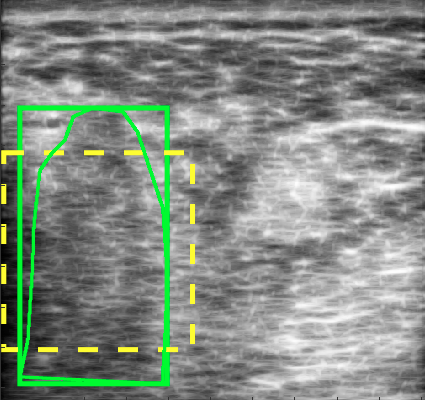}
\caption{Sample results with ground-truth contour and its bounding box (\emph{solid}) versus estimated (\emph{dashed}) ROI. }
\label{fig:sampleResults}
\end{center}
\end{figure}

\begin{algorithm}[h!]
\caption{Proposed Approach}
\begin{algorithmic}[1]
\label{alg:BarcodeROIEtimation}
\STATE \textbf{\% Index available images}
\STATE Read all images $I_1,I_2,\dots,I_N$ and their ground-truths $G_1,G_2,\dots,G_N$
\STATE Set counter $i=1$
\WHILE{$i \leq N$}
	\STATE Calculate a bounding box $\mathbf{b}_i=(x_s,y_s;x_e,y_e)$ around tumour marked in $G_i$
	\STATE Generate a global Radon barcode $\mathbf{r}_i^{\textup{G}}$ for $I_i$
	\STATE Generate a local Radon barcode $\mathbf{r}_i^{\textup{ROI}}$ for cropped ROI in $I_i$ determined by $\mathbf{b}_i$
	\STATE $i = i + 1$
\ENDWHILE
\STATE  \textbf{\% Process new images}
\STATE Read a query (new) image $I_q$
\STATE Get the centre of the tumour $(x_c,y_c)$ (user input)
\STATE Construct a fixed-size bounding box $\mathbf{b}_q$ around the centre ($\Delta\!=\!0.25$, $R,C$ image dimensions) with $x_s^q = \max(1,x_c-\Delta C)$, $y_s^q= \max(1,y_c-\Delta R)$, $x_e^q=\min(C,2 \Delta C)$, and $y_e^q= \min(R, 2 \Delta R)$;
\STATE Generate a global Radon barcode $\mathbf{r}_q^{\textup{G}}$ for $I_q$
\STATE Generate a local Radon barcode $\mathbf{r}_q^{\textup{ROI}}$ for cropped ROI in $I_q$ determined by $\mathbf{b}_q$
\STATE Set counter $i=1$
\WHILE{$i \leq N$}
	\STATE Calculate the global Hamming distance: \\$\mathbf{d}_{\textup{H}}^{\textup{G}}(i)\leftarrow \textrm{XOR}(\mathbf{b}_i^{\textup{G}},\mathbf{b}_q)$
	\STATE Calculate the local Hamming distance: \\$\mathbf{d}_{\textup{H}}^{\textup{ROI}}(i)\leftarrow \textrm{XOR}(\mathbf{b}_i^{\textup{ROI}},\mathbf{b}_q)$	
	\STATE $\mathbf{d}_{\textup{total}}(i) = \mathbf{d}_{\textup{H}}^{\textup{G}}(i) + \mathbf{d}_{\textup{H}}^{\textup{ROI}}(i)$
	\STATE $i = i + 1$
\ENDWHILE
\STATE Sort total differences $\mathbf{d}$ in decreasing order: \\$\mathbf{v}\leftarrow \textrm{SORT}(\mathbf{d})$
\STATE Estimate the bounding box $\mathbf{b}_q$ using Eqs.~(\ref{eq:estimate}) and~(\ref{eq:weights})
\end{algorithmic}
\end{algorithm}

\section{Leave-One-Out Validation}
The Radon barcodes were generated using the Matlab code available on-line\footnote{Matlab code available on-line: http://tizhoosh.uwaterloo.ca/}. Input images were resized to $128\times 128$ pixels and 64 projections were used resulting in a global barcode of 8192 bits. Training ROIs were bounding boxes around the ground-truth. Each cropped ROI was resized to $64\times 64$ pixels with 32 projections resulting in an ROI barcode of 2048 bits. There were 33 images available in total with an expert marking. A leave-one-out validation test was employed to evaluate the performance of the proposed method. The experiments generated 33 instance measurements amounting to a total accuracy of $81\% \pm 14\%$ (see Fig.~\ref{fig:sampleResults} for sample results). To simulate the user mouse click in the centre of the tumour, the center of the ground-truth was used. A fixed size ROI (quarter of the image dimensions) was then constructed in order to extract the ROI barcode during testing. Variations in the centre locations did not seem to have a major impact as long as they were within the close vicinity of the actual centre point. Future work will include using the estimated ROIs to characterize tumours textures for cancer treatment prediction. Besides, further improvements are necessary to increase the estimation accuracy. And finally, more data should be acquired to more accurately validate the performance of the proposed methods.

\section {Conclusions}
In this work, preliminary results were presented on ROI estimation in ultrasound B-mode images of locally advanced breast cancer. It was demonstrated that using global and local Radon barcodes could provide a fast semi-automated rough localization of tumours in modalities like ultrasound where noise and the inherent vagueness of tumour borders make contour extraction extremely challenging.

\bibliographystyle{IEEEbib}
\bibliography{refs}

\end{document}